\documentclass[nouppercase]{kaia}
\usepackage[english]{babel}
\usepackage{hyperref}
\addto\extrasenglish{%

}
\usepackage{amsmath}
\title{City-Scale Visual Place Recognition with Deep Local Features Based on Multi-Scale Ordered VLAD Pooling}

\author{Duc Canh Le}{a}
\author{Chan-Hyun Youn}{a}
\affiliation{Department of Electrical Engineering, KAIST, Daejeon, Korea \{canhld,chyoun\}@ kaist.ac.kr}{a}

\begin{document}

\maketitle
\begin{abstract}
Visual place recognition is the task of recognizing a place depicted in an image based on its pure visual appearance without metadata. In visual place recognition, the challenges lie upon not only the changes in lighting conditions, camera viewpoint, and scale, but also the characteristic of scene level images and the distinct features of the area. To resolve these challenges, one must consider both the local discriminativeness and the global semantic context of images. On the other hand, the diversity of the datasets is also particularly important to develop more general models and advance the progress of the field. In this paper, we present a fully-automated system for place recognition at a city-scale based on content-based image retrieval. Our main contributions to the community lie in three aspects. Firstly, we take a comprehensive analysis of visual place recognition and sketch out the unique challenges of the task compared to general image retrieval tasks. Next, we propose yet a simple pooling approach on top of convolutional neural network activations to embed the spatial information into the image representation vector. Finally, we introduce new datasets for place recognition, which are particularly essential for application-based research. Furthermore, throughout extensive experiments, various issues in both image retrieval and place recognition are analyzed and discussed to give some insights for improving the performance of retrieval models in reality.
\end{abstract}

\section{Introduction}
Visual place recognition has received a significant amount of attention in the past few years both in computer vision and robotics communities , motivated by applications in autonomous driving, augmented reality and simultaneous localization and mapping (SLAM) \cite{Arandjelovic2018,Torii2018}. Generally, there are two major approaches for visual place recognition: image retrieval-based and large-scale 3D model-based. 

In the large-scale 3D model approach, the scenes are represented via 3D models with image descriptors attached to 3D points and the localization task is cast as 2D-to-3D image registration where the full 6 DOF camera poses are recovered from the models \cite{Torii2019}. The method has the advantage of high accuracy, nevertheless, building and maintaining a large-scale 3D model is extremely expensive. On the other hand, the image retrieval based methods approximate the location of a query image with locations of similar images in a large-scale geo-tagged database. Compared to 3D-based methods, maintaining a database of geo-tagged images is much easier, therefore, the approach has the advantage of scalability. Moreover, it has been shown in \cite{Torii2019} that retrieval-based method can also recover both position and full camera poses of the query image if sufficient database images are correctly retrieved. 

In general image retrieval, the goal is to find the image with the most similar visual appearance to the instance depicted in the query. Modern retrieval systems index an image to a compact vector, which captures both the local and the semantic visual appearance of the image. By measuring the distance between these vectors, we can compute the similarity between the corresponding images. The fundamental issue in image retrieval is how to create a canonical form that efficiently unify similar images and distinguish those are different.  

Visual place recognition exposes unique challenges besides to those existing in general image retrieval. Nowadays, challenges of visual place recognition come from not only the visual differences between query and database images but also from the overlapping between irrelevant images, and the different appearance of the same place among images \cite{Kim2017}. For example, in city scenes, common dynamic objects such as cars, humans, bikes, trains appear in most of the images; or a picture of same place may shift depending on the camera model, the viewpoint, the time in the day or the season it was taken.

To address these issues, we propose a pooling scheme on top of convoluional neural network (CNN) activations that produce an effective vector-form of scene images. Our method takes advantage of local features in CNN, pool them with a statistical model to achieve the canonical form, and finally exploit the multi-scale transform to retain the spatial details. While the final vector may include residual information, the experiment show that it effectively eliminates the unique issues in visual place recognition. 

In addition, one of the challenges in the research of visual place recognition in the early days was how to collect enough geo-tagged images to build the database. In recent years, map providers such as Google or Naver introduce the street-view feature in their map platform, which allows users to view the street-level spherical panorama image in most urban areas in the world. Subsequently, several city-scale databases have been constructed \cite{Torii2015,Arandjelovic2018,Schindler2007}. Nevertheless, our significant concern when starting this research is whether or not the characteristics of cities affects the recognition methods. Encouraged by this question in mind, we generate the dataset of our city, Daejeon, with database images from Google Streetview and query images were taken by phones. We later show that our concern does make sense, as the state-of-the-art methods on existing datasets perform worse in our dataset.

The paper is organized as follow: In \autoref{sec:bgr} we describe in detail the challenges of visual place recognition and discuss previous works. In \autoref{sec:med}, we present our pooling scheme. In \autoref{sec:exp}, we first describe our newly collected dataset and then show demonstrate the benefits of proposed method over prior works. Finally, the conclusion and discussion on the future of visual place recognition are given in \autoref{sec:ccd}.

\section{Background and Related Works} \label{sec:bgr}

Content-based image retrieval (CBIR) has been a core problem in the multimedia field over two decades. Modern CBIR systems usually consist of two-stage: offline stage and online stage. In the offline stage, a visual database is constructed by crawling images from various sources to create the plaint-image pool, and indexing images to database for efficient searching. In the online stage, the query image is given and similar images in the database are retrieved by scoring the images in the database and optionally re-rank the top images with the highest scores by geometric verification \cite{Fischler1981}.

Nowadays, the challenge of CBIR lies in developing an efficient representation that is robust to different appearances of an image due to occlusions or changes in illumination, view-point, and scales. To achieve the invariant, local features is extracted from images. A local features describe a very small patch of the image (e.g. $16\times16$ pixels) and is robust to change in illumination and scale. Notable local features are SIFT \cite{Lowe2004}, SURF \cite{Bay2006}, BRIEF \cite{Calonder2010}; and recently off-the-shell CNN activations shows outstanding performance when used as local features \cite{Razavian2014}.

While local features is good for matching images, it is not suitable for image retrieval due to extra cost of point-to-point matching. To solve this problem, methods are developed to aggregate local features into a compact global feature. Most of the successful feature aggregation model are statistical-based, namely bag of visual words (BoVW) \cite{Sivic2003}, Fisher vector (FV) \cite{Perronnin2010} and vector of locally aggregated descriptors (VLAD) \cite{Jegou2010}. These models capture the distribution of local features over a codebook and use it as the global representation of images.

In recent years, end-to-end deep learning model introduces simple aggregation strategy on CNN feature and achieve decent global features by jointly learn both extractions and aggregation. \cite{Babenko2014} trained the CNN with classification loss and used the pooled vector at the last fully-connected layer as the global representation of the images. The authors of \cite{Radenovic2019,Tolias2015,He2015} replaced the fully-connected layer with pooling layer and trained the network with the triplet-ranking loss to produce a global image representation that is suitable for image retrieval. Remarkably, \cite{Arandjelovic2018} proposed the NetVLAD layer, which pools the features in VLAD manner and ready to plug into any CNN architecture. For visual place recognition, the authors showed that the whole network can be weakly supervised trained with triplet ranking loss and noisy data with only GPS geo-tagged.

When applied to visual place recognition, CBIR reveals unique challenges due to the characteristic of urban environment. The issue come from not only the visual differences between query and database images but also from the overlapping between irrelevant images, and the different appearance of the same place among images \cite{Kim2015}. For example, in city scenes, common dynamic objects such as cars, humans, bikes, trains appear in most of the images; or a picture of same place may shift depending on the camera model, the viewpoint, the time in the day or the season it was taken. Many efforts have been made in the recent decade to solve these challenges with attentive features model, on which the features extractor is trained to detect key-points at the salient areas of the images, i.e. static objects such as walls or buildings. However, in \cite{Kim2017,Torii2015,Kim2015} and in this paper, we show that the attentive model itself is not strong enough for better place recognition. The underlying reasons lie in two issues: the repetitive structures , and the reflection phenomenon.

\subsection{Repetitive Structures} \label{subsec:repe}

Repetitive structures may refer to the common patterns that appear frequently in many different geographical places and are easily matched to other instances of the same type. The examples of repetitive structures in the city are generic windows, building fences or trees.There are two key reasons why the repetitive pattern is problematic in the retrieval algorithm. First, the features from these patterns dominate in the image and therefore, degrade the contributions of other important features. For example, in the BoVW model, the final representation vector is actually the histogram of features over the pre-trained codebook. Then if two images are completely different but they are dominant by the same repetitive structure, the BoVW model is likely to give them a relatively high similarity score. The same issue can be found in VLAD model, as they are both pure statistical-based. The second reasons are the representation of images in the database usually ignores the spatial relations between features. This can be improved by post-processing methods such as spatial verification with RANSAC \cite{Fischler1981} with the penalty on computational complexity. However, post-processing is not always effective because sometimes the correct images cannot even reach to spatial verification step.

\begin{figure}[t]
\begin{center}
\includegraphics[width=0.46\textwidth]{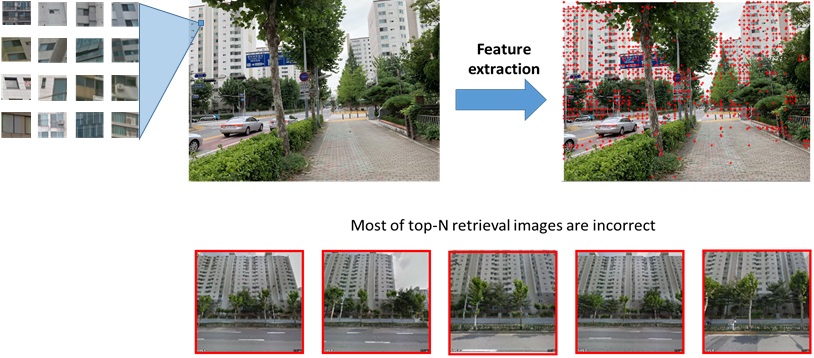}
\end{center}
   \caption{Typical building facades in our database that create difficulty to distinguish buildings; many features extracted from the image belong to repetitive patterns, thus, the retrieval system return many false-positive images}
   \label{fig:repetitives_structure}
\end{figure}

There are efforts to reduce the negative impact of repetitive structures on retrieval algorithm \cite{Kim2015,Kim2017,Torii2015,Sivic2003}. Especially, the author of BoVW \cite{Sivic2003} proposed to use term frequency – inverted documents frequency (TF-IDF) weighting scheme on the visual words of the codebook, in which the more frequently visual words are down-weighted. Formally, suppose there is a codebook of $V$ visual words, then each image is represented by a vector:
\begin{equation}
v_d = (t_1,t_2,…,t_V)^T                          
\end{equation}

\noindent of weighted visual word frequencies with components:

\begin{equation}
t_i = \frac{n_{id}}{n_d}\times\log\frac{N}{N_i} 
\end{equation}

Where $n_{id}$ is the number of occurrences of visual words $i$ in image d, $n_d$ is the total number of visual words in the image $d$, $N_i$ is the number of images containing term $i$, and $N$ is the number of the whole database. TF-IDF degrades the impact of repetitive patterns with the assumption that they are common and dominated in the whole database. However, this assumption does not always hold owing to two factors: (1) repetitive patterns may appear in many images but not major of the database, and (2) the features from these patterns are not necessarily identical. Therefore, [49] and [2] suggested explicitly detecting and down-weighting the features from repetitive structures (that called burstiness features). Recently, \cite{Kim2015,Kim2017} introduces a method to learn the good features for visual place recognition and the model implicitly learns to down weight the features from the repetitive structures.

All of the prior works are statistical-inspired and aim to break the domination of burstiness features. On the other hand, our approach attempts to embed the spatial information among features, which includes the burstiness features, to the final representation. We later show that our method and prior methods are complementary and one can combine both to further improve the efficiency of place recognition algorithms.

\begin{figure}[t]
\begin{center}
\includegraphics[width=0.46\textwidth]{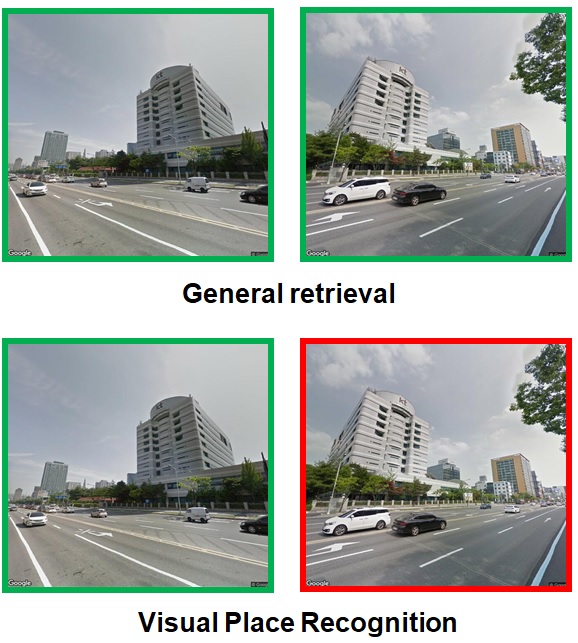}
\end{center}
   \caption{Two image of the same building from different perspectives; in general retrieval, they are considered similar, but in visual place recognition, they are false positives}
   \label{fig:reflex}
\end{figure}

\subsection{Reflection Phenomenon} \label{subsec:reflex}

Reflection phenomenon usually happens when images contain a large building. In some general image retrieval tasks, e.g. product retrieval, landmark recognition, and visual search, etc., the global context information is not an essential concern. For example, if one queries the visual search engine with an image of Eiffel tower, then all retrieved images should merely contain the tower no matter the distance or the perspective of these images to the tower. Similarly, if one wants to search the information of a product from an image, he only cares about the product itself but not the semantic context of the image. However, in visual place recognition, the ultimate goal is to detect the location of the query image, therefore, the semantic context information of images does make sense and should be thoughtfully concerned.

The reflection phenomenon is very unique to visual place recognition compared to other issues. Similar to the repetitive phenomenon, it can also be improved with spatial verification with the penalty on computational cost and processing time. Nevertheless, in spatial verification, we verify the relation among all features, which is quite not necessary in this case. Indeed, all we need is a semantic representation of images to degrade the negative impact of both repetitive structures and reflection phenomena. In the next Section, we present yet a simple technique to generate a semantic representation of an image with multi-scales ordered VLAD pooling of the image deep local features.

\begin{figure*}[t]
\begin{center}
\includegraphics[width=0.95\textwidth]{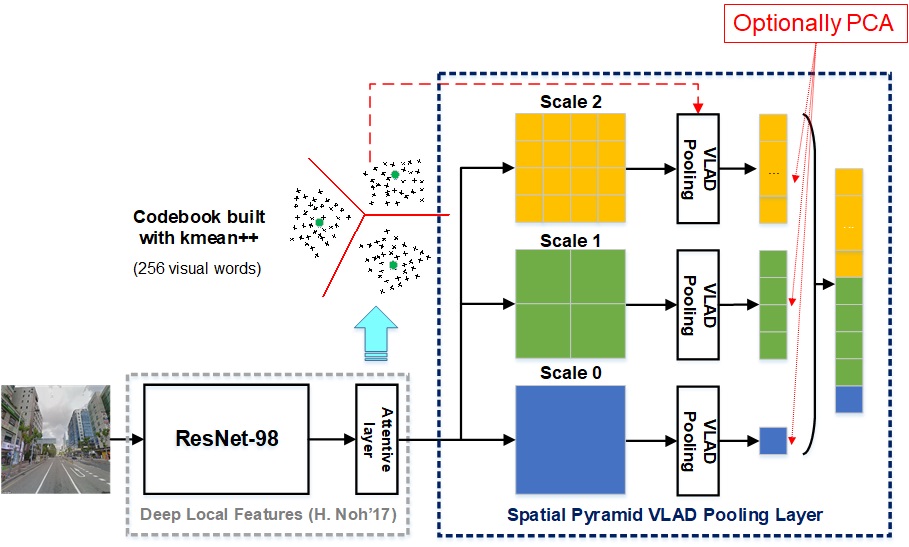}
\end{center}
   \caption{The overall framework of our Spatial Pyramid VLAD Pooling approach for visual place recognition; in sort, we pool the CNN features with VLAD in an ordered manner}
   \label{fig:framework}
\end{figure*}

\section{Methodology} \label{sec:med}

\subsection{Pooling Features on the Top of CNN Activations} \label{subsec: cnnpooling}

In recent years, CNN has shown breakthrough performance on many computer vision tasks compared to traditional models. Many research describes CNN as a “brain-inspired” computing model, however, in computer vision, CNN yet still follows the conventional pipeline of computer vision and pattern recognition: extracting image features and learning the patterns on top of these features. The underlying idea of CNN compared to the traditional machine learning approach is: in traditional machine learning, the features are extracted manually and the classification model is learnable, but in CNN, both the features and the classification model are immensely learnable. Therefore, the last fully connected layer can be replaced with other machine learning models to perform tasks other than image classification or improve the accuracy of the model \cite{Razavian2014}.

Our paper focuses on Streetview images, where the human-made constructions and their locations are significant to recognize the scene. In scene level images, the image usually consists of many objects on a background. The background occupies a large portion of the image but usually has a uniform appearance and does not tell us much about the uniqueness of the image. On the other hand, foreground objects, even engage relatively few pixels in the image, contains far more useful information. In scene-level images, there is no priority among objects, and the spatial relations among objects are essential for the semantically understanding the image. For the visual place recognition task, we expect the retrieved images should not only contain the same buildings but also have a similar viewpoint with the query image. To adding the spatial information to the image, we exploit the ordered-pooling approach, where the images are orderly compared at multi-scales. We refer our method as Spatial Pyramid VLAD Pooling (SPVP).

\subsection{Spatial Pyramid VLAD Pooling} \label{spvp}

Our work also relies on reusing pre-trained CNN features as the off-the-shell local features, but instead of pooling them in a trivial manner as previous works, we aim to develop a more sophisticated aggregating method. At first glance, one can use BoVW \cite{Sivic2003} or VLAD \cite{Jegou2010}, which has been already thoroughly studied and shown sufficient performance with hand-crafted local features. However, these statistical pooling methods suffer from missing spatial information of the images. Inspired by the Spatial Pyramid Pooling Network (SPP-net) \cite{He2015} and R-MAC \cite{Tolias2015} which extracts the image features at a single scale but pool them over regions of increasing scales, we propose to pool the CNN activations in VLAD manner at multi-scales. The detail of our framework is shown in \autoref{fig:framework}.

Our framework is built on top of Deep Local Features (DELF) \cite{Noh2017} with the backbone network is Resnet50 \cite{He2016} following by an attentive layer. In DELF, the $2048-D$ activations are taken at the output of the $conv4\_x$ block of Resnet50 \cite{He2016} and fed to the attentive layer to select semantically meaningful features for place recognition. The attentive model can be trained implicitly with classification loss. The selected features are L2-normalized, and then their dimension is reduced to 40 with principal component analysis, and finally L2-normalized again. We use both the pre-trained FCN model on the Google Landmark v2 dataset and the PCA model from the authors1 to extract the features from the images and from there develop our representation technique. In our framework, the default image size is $640\times640$ and the DELF outputs a sparse $640\times640\times40$ features map. Our representation has total of three scales, corresponding to activations of original $640\times640$ image and $320\times320$ and $160\times160$ patches. Note that we only pass the image through the extractor network once and then build the pyramid on the final activations.

Next, we need to pool the activations of the image and its patches to summarize the representation by a single feature vector of reasonable dimension. For this, we adopt VLAD \cite{Jegou2010} instead of mean pooling \cite{Radenovic2019} or max-pooling \cite{Tolias2015}. We randomly select 10 million features to build a codebook of 256 visual words with k-mean, and at each level, we pool the features of each patch in VLAD manner with respect to the codebook. Each DELF feature is $40-D$, results in a $10240-D$ VLAD vector for each patch, which is too high. Therefore, we train the PCA on top of all VLAD vectors to reduce the dimensionality of these vectors to either $256$, $512$, $1024$ or $2048$.

Intuitively, one can describe our methods as multi-patches pooling in the sense that we divide the image into multiple patches and simultaneously compare image-to-image and patch-to-patch. This strategy indeed works in visual place recognition by following justifications: (1) the patch-to-patch comparison can preserve the spatial relations among objects in the images and ensure that if two images are matched by patch-to-patch comparison, they are semantically similar; and (2) even if in the database, there are no image with perfectly view-point that can be matched with patch-to-patch, the global image-to-image comparison yet still works and provide us the most similar images in the database. In other words, our pooling method always guarantees the best match images are retrieved for a given query.

\textbf{Relation to previous studies: }Our approach is inspired by SPP-net \cite{He2015} and R-MAC \cite{Tolias2015}, nevertheless, these approaches are not specialized for place recognition and their pooling operator are trivial (max-pooling). On the other hand, our pooling method is designed particularly for place recognition and we pool the CNN features with VLAD. The most related method to ours is the one of Gong et. al. \cite{Gong2014}, which proposes to pool the CNN features with VLAD at multi-scale in an orderless manner. We show the difference between ordered and orderless pooling in Figure 3-3. In short, the orderless pooling summarizes the patch-level VLAD vector while the ordered pooling concatenates them. Nevertheless, we argue that their work is not applied well to visual place recognition, because orderless pooling is good to handle object at different scales but do not preserve the spatial information of the image. The second key difference is in their work, the representation vector at the first scale is corresponding to $4096-D$ CNN activation for the entire $256\times256$ image, and the VLAD is only applied at higher scales of the pyramid. Meanwhile, we use VLAD on all scales.

\begin{figure}[t]
\begin{center}
\includegraphics[width=0.46\textwidth]{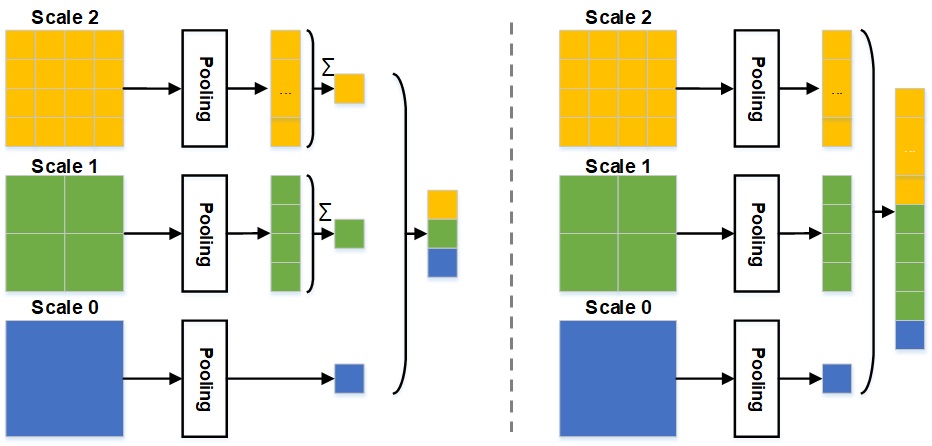}
\end{center}
   \caption{Our ordered pooling (Left) vs Gong. et. al. \cite{Gong2014} orderless pooling (Right); even achieving more compact vector size, ordredless pooling suffers from missing of spatial information}
   \label{fig:ordered_vs_orderless}
\end{figure}

\section{Experiments} \label{sec:exp}

\subsection{Dataset and Evaluation Metric} \label{subsec:dataset}

We exploit the Google Streetview Imagery to create the database of the city. Specifically, we select the areas of interest in the city and divide them into a $10m\times10m$ grid and the street-view images are taken at each point on the grid. At each location, we capture 8-perspective images of $640\times640$ pixels with a pitch direction $10^o$ and the following 8 yaw directions $[0^o, 45^o,…, 360^o]$. We use that GPS location as the geo-tag in the database.

\begin{figure}[t]
\begin{center}
\includegraphics[width=0.46\textwidth]{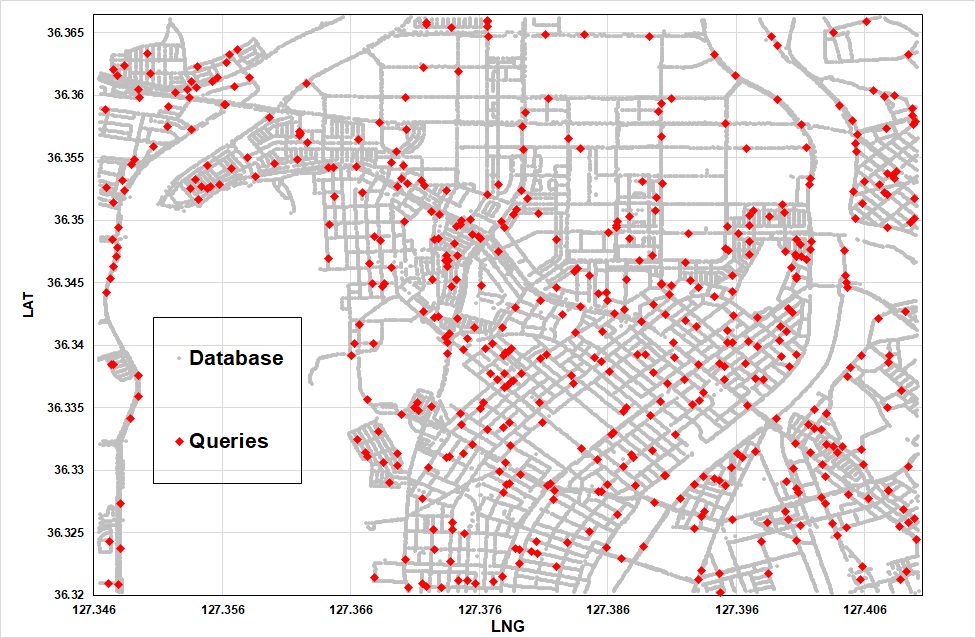}
\end{center}
   \caption{Locations of the database (blue) and query (red) images in the dataset; the dataset includes 259.064 database images and the query set with 4.000 images, distributed over 500 locations in the area}
   \label{fig:dataset}
\end{figure}

After the crawling process, there is a total of $86.885$ locations are captured, results in a total of $695.080$ street-level images with GPS tags. We use these images to create the experiment datasets with $263.064$ images, covers $40 km^2$ of the city's urban area. We split those images into two sets: the database with $259.064$ images and the query set with $4.000$ images, equally distributed over $500$ locations in the area. The visualized locations the dataset are shown in Figure 4-1 , respectively. In the experiment, we use GPS as the growth truth location for queries. For each image in the query set, correct locations of the image are all street-view points with the distance to the image’s GPS location less than $D$ meters with $D$ is from $10$ to $50$. 

For each query, we retrieve top-N most similar images from the database and say that this query is correctly recognized at N if at least one in top-N retrieved images is at the correct point. Then we calculate the percentage of correctly recognized queries ($Recall@N$) as well as the percentage of correct images in the retrieved result ($Precision@N$) with respect to the values of N and methods. Formally, let \({Q_i}:i = (1-M)\) is the set of queries, \(R_{k}^{(Q_j)}:k= (1-K)\) is the retrieved result vector of the query $Q_j$ with $R_{k}^{(Q_j)}$ is either 1 for correct images or 0 for wrong images. Then with \( N \leq K \), the Recall@N is calculated by:

\begin{equation}
Recall@N = \frac{\sum_{i=1}^{M} ({OR}_{k=1}^{N} R_{k}^{Q_i})}{M}
\end{equation}

\noindent and the Precision@N is calculated as:

\begin{equation}
Recall@N = \frac{\sum_{i=1}^{M} (\sum_{k=1}^{N} R_{k}^{Q_i})}{M}
\end{equation}

\subsection{Quantitative Experiment}
\begin{figure}[t]
\begin{center}
\includegraphics[width=0.46\textwidth]{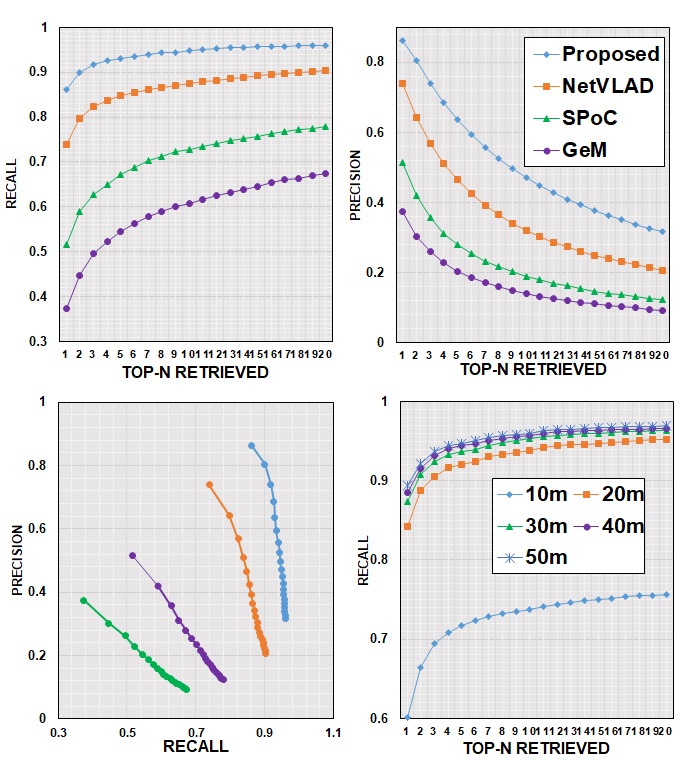}
\end{center}
   \caption{SPVP vs. STOA – recall and precision vs number of retrieved images; SPVP outperforms all current state-of-the-art pooling methods by a large margin; On the last figure, SPVP also perform reasonably with different error threshold}
   \label{fig:quantex}
\end{figure}
\begin{figure*}[t]
\begin{center}
\includegraphics[width=0.96\textwidth,height=6cm]{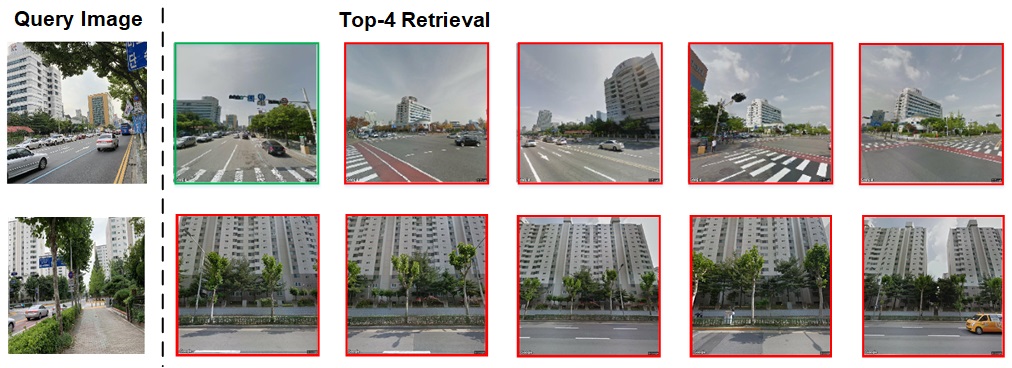}
\end{center}
   \caption{Visual failure cases of NetVLAD \cite{Arandjelovic2018} due to missing of semantical information of the baseline model; the green bound denotes the correct image and red bound denotes the wrong image; in the above query, the model suffers from the “reflection effects”, and in the below query, the model suffers from the repetitive structures}
   \label{fig:qual_ref}
\end{figure*}
\begin{figure*}[!ht]
\begin{center}
\includegraphics[width=0.96\textwidth,height=6cm]{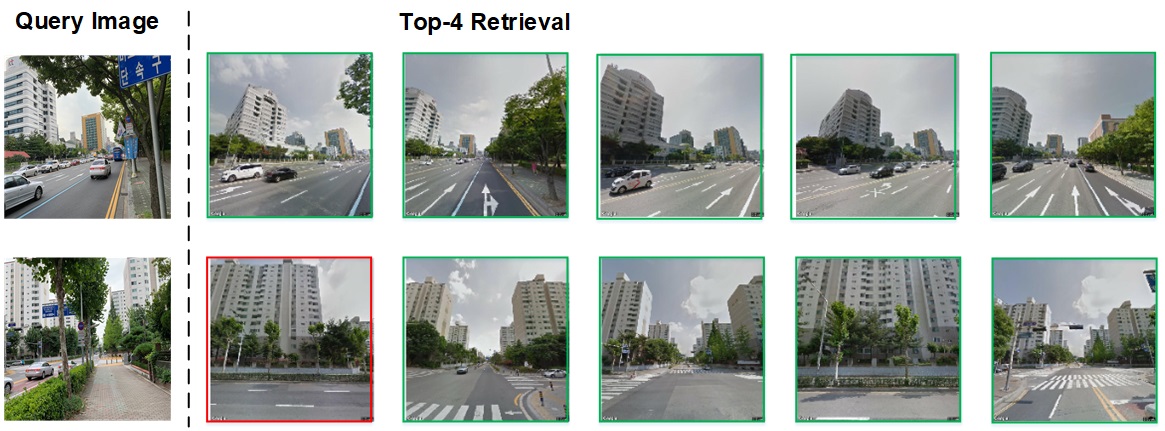}
\end{center}
   \caption{SPVP can overcome the repetitive structures as well as provide more semantic representation of images with most of retrieved images are at the correct locations; moreover, even SPVP is designed in mind to deal with the repetitive patterns and spatial information loss, it still performs well on ordinary queries}
   \label{fig:qual_pro}
\end{figure*}

We demonstrate our proposed SPVP scheme on our collected dataset. We also employ popular pooling methods on top of CNN activations and state-of-the-art end-to-end learning method to compare with our pooling scheme, namely max-pooling (MAC) \cite{Tolias2015}, mean (SPoC) and generalized mean (GeM) \cite{Radenovic2018} pooling, and NetVLAD \cite{Arandjelovic2018}. We report the recall and the precision curve of all approaches in \autoref{fig:quantex}. 

On our dataset, SPVP outperforms all previous pooling methods, including the state-of-the-art NetVLAD, by a large margin. SPVP achieves top-1 recall of $86\%$ and top-20 of $96\%$, which is leading over NetVLAD ($74\%$ and $90\%$), SPoC ($49\%$ and $79\%$), and GeM ($37\%$ and $68\%$). Remarkably, MAC performs extremely poor on our dataset with top-1 and top-20 recall are both smaller than $1\%$ (that we do not include it in the charts!). Similarly, SPVP exclusively reaches higher performance than other approaches in both precision and precision-recall experiment. For example, at the $40\%$ precision, GeM, SPoC, and NetVLAD attain recalls of $40\%$, $60\%$, and $85\%$, while SPVP reaches $95\%$ recall at the same precision. Nevertheless, the improvement of our model comes with the cost of memory: the length of representation vectors in our approach is 51.200, much higher than SPoC (40), GeM (40), and NetVLAD (4.096). We believe this drawback can be resolved when combining our approach with dimension reduction techniques like PCA.

In the previous experiments, we set the distance error threshold D=25m. However, in practice, this value can vary depending on the application. Therefore, we examine our method with different distance thresholds from $10m$ to $50m$ and step size of $10m$, and report the recall value at each distance threshold in \autoref{fig:quantex}. The result indicates that the recall is significantly reduced with $D = 10m$ with the value at top-1 and top-20 is $60\%$ and $75\%$. Nevertheless, we argue that this is normal because of two reasons: (1) we collect the database with a $10m\times10m$ grid, therefore $D = 10m$ is smaller than the resolution of the database; and (2) even the GPS (which we use as the growth truth in our experiment) itself has error of 10m - 15m. When $D > 10m$, our approach consistently produces good performance with recall values from $85\%$ to $95\%$.

\subsection{Qualitative Experiment}

When we did experiment on some specific queries, we demonstrate that SPVP can overcome the repetitive structures as well as provide more semantic representation of images. Figure \autoref{fig:qual_ref} shows the queries and its top-4 retrieval result with NetVLAD from the database with the green bound denotes the correct image and red bound denotes the wrong image. On the same queries, as shown in \autoref{fig:qual_pro}, SPVP does not suffer from repetitive structures as well as the view-point changes and most of the retrieved images are at correct locations

\begin{figure}[t]
\begin{center}
\includegraphics[width=0.46\textwidth]{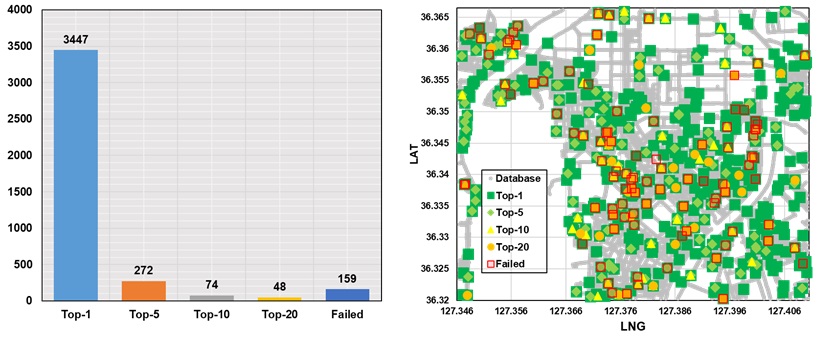}
\end{center}
   \caption{Statistic of recognition result and heatmap of the queries $86\%$ of queries are successfully recognized at top-1 and $4\%$ are totally failed under top-20; both good queries and failed queries are equally likely over the whole map without any bias}
   \label{fig:heatmap}
\end{figure}

\subsection{Statiscal Analysis of the Dataset}

Finally, we report the statistical of the experiment on the dataset and show the result in \autoref{fig:heatmap}. Among 4000 queries, $3447$ queries ($86\%$) are successful recognized at top-1, $272$ queries ($6.8\%$) at top-5, $47$ queries ($1.1\%$)  top-10, and $48$ queries ($1.1\%$) at top-20. The remaining $159$ queries ($4\%$) are totally failed under top-20. We show the heat-map of the queries in \autoref{fig:heatmap} with the color of green indicates good recognition performance queries (successful recognized at top-1) and red for failed queries. We can observe that the distribution of both good queries and failed queries are equally over the whole map. Therefore, we conclude that our large scale dataset is fair enough to be utilized in other experiments out of this thesis.

\section{Conclusion and Discussion} \label{sec:ccd}

\subsection{Concluding Remark}
In this thesis, we present a fully-automated system for place recognition at the city-scale based on content-based image retrieval. Our contributions for the community lie in three aspects: (1) we take a comprehensive analysis of visual place recognition and sketch out the unique challenges of the task; (2) we propose yet a simple pooling approach on top of CNN features to embed the spatial information into the image representation vector; and (3) we introduce new datasets for place recognition, which are particularly essential for application-based research.

\subsection{Discussion}

\subsubsection{Visual place recognition in reality: }
Visual place recognition has many potential applications in a wide range of services. In robotics, visual simultaneous localization and mapping (SLAM) have become an emerging field of research in recent years. SLAM is the task of constructing and updating the map of an environment while simultaneously maintaining the location of the agent within it. Visual SLAM does the task by using visual information only, therefore, techniques of visual place recognition can be easily adopted in visual SLAM. On the other hand, visual place recognition has been utilized in commercial photo collection services. Google Photos is the cloud-based service that allows users storing and organizing their photos on the cloud. In Google Photos, the location of the image can be either the GPS or predicted based on its visual appearance when GPS is not available. In the future, we believe that visual place recognition can be useful in many other emerging platforms such as smart city and smart building (indoor localization).

\subsubsection{Deep learning and image retrieval: }
Deep learning and image retrieval: In \autoref{sec:bgr}, we have shown that deep learning is radically changing the general framework of image retrieval, nevertheless, how much should deep learning be adopted in the framework, is somewhat an interesting question. In \cite{Arandjelovic2018}, the authors state that “the core behind the idea that makes the success of deep learning is end-to-end learning”, and suggest that the general framework can be replaced with single-pass deep learning models. However, in terms of representation learning, the representation from single-pass models is not general and coupled to one category of image retrieval. On the other hand, in the traditional framework, we learn the general image representation first and deploy the task-specific models on top of this representation, therefore, we can exploit more sophisticated techniques tailored with the task to improve the retrieval performance, e.g. spatial information embedding or exploiting the side features. Another major disadvantage of the single-pass models is they require complex training datasets with strong supervision.

\subsubsection{Legacy of the datasets: }In this thesis, we build our dataset with Google Streetview Imagery, following pioneers at \cite{Torii2015,Arandjelovic2018,Schindler2007}. However, the newest Google Maps Platform Terms of Service states that
\begin{quote}
    “...the customer will not extract, export, or otherwise scrape Google Maps Content for use outside the Services. For example, Customer will not: (i) pre-fetch, index, store, reshare, or rehost Google Maps Content outside the services; (ii) bulk download Google Maps tiles, Street View images, geocodes, directions, distance matrix results, roads information, place information, elevation values, and time zone details; (iii) copy and save business names, addresses, or user reviews; or (iv) use Google Maps Content with text-to-speech services”
\end{quote}

Therefore, even we do not use the Google Map API, our datasets may still violate the Googe Terms of Uses. Nevertheless, there are open map platforms that support building the Streetview database and therefore, can be utilized in the future for constructing datasets for place recognition.

\section{Acknowledgement} \label{ack}

\bibliography{references}

\newpage
\onecolumn
	\begin{center}
		\section*{Summary of this paper}	
		\setcounter{subsection}{0}
	\end{center}
\noindent\fbox{
    \parbox{\textwidth}{
    	\begin{quote}
		\subsection{Problem Setup}
		We apply content based image retrieval to place recognition in urban area. Due to the characteristic of urban environment such as repetitive structures and reflection phenomenon, general global feature introduce many false-positives. In this paper, we aim to develop an image representation that effectively capture the essential information of city-view images.
		\vspace{0.5\baselineskip}
		\subsection{Novelty}
		We propose a pooling scheme on top of convoluional neural network (CNN) activations that produce an effective vector-form of scene images. Our method takes advantage of local features in CNN, pool them with a statistical model to achieve the canonical form, and finally exploit the multi-scale transform to retain the spatial details.
		
		We introduce new dataset for place recognition. One of our significant concern when starting this research is whether or not the characteristics of cities affects the recognition methods. Encouraged by this question in mind, we generate the dataset of our city, Daejeon, with database images from Google Streetview. We later show that our concern does make sense, as the state-of-the-art methods on existing datasets perform worse in our dataset.
		\vspace{0.5\baselineskip}
		\subsection{Algorithms}
		We fine-tune the Resnet98 with Google Landmark dataset and use the output feature-map of the last CNN block as the local features. Then we aggregate the local features with vector of locally aggregated descriptors (VLAD) and finally, pool all of them in ordered manner to create the final representation of the image.
		\vspace{0.5\baselineskip}
		\subsection{Experiments}
		We demonstrate our proposed method on our collected dataset. We also employ popular pooling methods on top of CNN activations and state-of-the-art end-to-end learning method to compare with our pooling scheme, namely max-pooling (MAC) \cite{Tolias2015}, mean (SPoC) and generalized mean (GeM) \cite{Radenovic2018} pooling, and NetVLAD \cite{Arandjelovic2018}
		
		We show that on our dataset, SPVP outperforms all previous pooling methods, including the state-of-the-art NetVLAD, by a large margin. 
		
		When we did experiment on some specific queries, we demonstrate that SPVP can overcome the repetitive structures as well as provide more semantic representation of images.
		\begin{center}
		\includegraphics[width=0.65\textwidth]{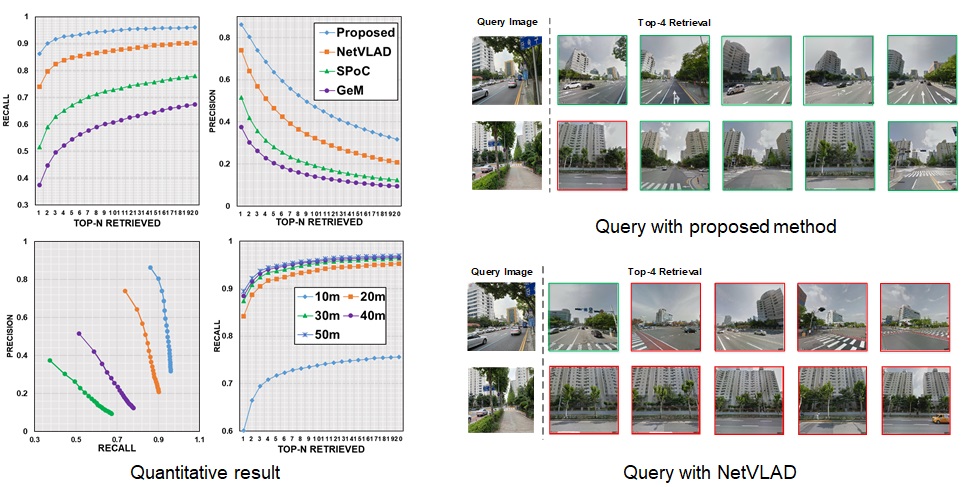}
		\end{center}
		\vspace{0.5\baselineskip}
		\end{quote}
    }
}
\thispagestyle{empty}
\end{document}